\documentclass{article} 
\usepackage{iclr2027_conference,times}

\usepackage{amsmath,amsfonts,bm}

\def\eqref#1{equation~\ref{#1}}

\def\1{\bm{1}}

\DeclareMathAlphabet{\mathsfit}{\encodingdefault}{\sfdefault}{m}{sl}
\SetMathAlphabet{\mathsfit}{bold}{\encodingdefault}{\sfdefault}{bx}{n}

\usepackage{hyperref}
\usepackage{url}
\usepackage{graphicx} 
\usepackage{amsmath,amssymb,bm}
\usepackage{booktabs} 
\usepackage{multirow}
\usepackage{etoolbox}
\usepackage{xspace}
\usepackage{xcolor}
\usepackage{tikz}
\usepackage{wrapfig}
\usepackage{caption}
\usepackage{adjustbox}
\usepackage{xcolor}
\usepackage{algorithm}
\usepackage{algorithmic}

\definecolor{nextscalebg}{HTML}{DCEAF7}
\definecolor{nextscalefg}{HTML}{245B87}

\DeclareRobustCommand{\Ntag}{%
    \begingroup
    \setlength{\fboxsep}{1.2pt}%
    \colorbox{nextscalebg}{%
        \textcolor{nextscalefg}{\strut\textbf{N}}%
    }%
    \endgroup
}
\title{MeshOctave: Vertex Split-and-Rewire Cascades for Native Mesh Generation}
\author{
    \normalfont
    \begin{tabular}{@{}l@{}}
    \small
    \textbf{Junkai Lin}$^{1,2}$
    \textbf{Tianhao Zhao}$^{1,2}$
    \textbf{Hang Long}$^{1,2}$
    \textbf{Huipeng Guo}$^{1}$
    \textbf{Jielei Zhang}$^{1}$
    \textbf{Youjia Zhang}$^{1,2}$
    \textbf{Jiale Xu}$^{2}$ \\
    \small
    \textbf{Wenbing Li}$^{1,2}$ 
    \textbf{Rendong Liang}$^{2}$
    \textbf{Jozef Hladký}$^{3}$
    \textbf{Matthias Nießner}$^{4}$
    \textbf{Yuanming Hu}$^{2}$ \hspace{0.1em}
    \textbf{Wei Yang}$^{1,\dagger}$ \\
    $^{1}$Huazhong University of Science and Technology \quad
    $^{2}$Meshy AI \\
    $^{3}$Independent Researcher \quad
    $^{4}$Technical University of Munich \\
    \\
    \multicolumn{1}{c}{%
    \href{https://maymhappy.github.io/MeshOctave/}{\textcolor{red}{\texttt{https://maymhappy.github.io/MeshOctave/}}}%
}
    \end{tabular}
}

\graphicspath{{figures/}}

\newcommand{\method}{\textsc{MeshOctave}\xspace}

\newcommand{\Mesh}{\mathsf{M}}

\newcommand{\Mintra}{M^{\mathrm{intra}}}
\newcommand{\Minter}{M^{\mathrm{inter}}}
\newcommand{\methodhead}[1]{%
    \begin{tabular}[c]{@{}c@{}}
        #1
    \end{tabular}%
}
\makeatletter
\patchcmd{\@maketitle}
  {\lhead{Published as a conference paper at ICLR 2027}}
  {\lhead{\textit{MeshOctave: Vertex Split-and-Rewire Cascades for Native Mesh Generation}}}
  {}{}
\makeatother

\iclrfinalcopy 
\begin{document}

\vspace*{-1.3cm}
\maketitle
\footnotetext[1]{
    \textbf{1,2} This work was done while interning at Meshy AI.
}
\footnotetext[2]{
    $\dagger$ Corresponding author:
    \href{mailto:weiyangcs@hust.edu.cn}{weiyangcs@hust.edu.cn}
}
\vspace{-0.7cm}
\begin{figure}[!h]
\centering
\includegraphics[width=\textwidth]{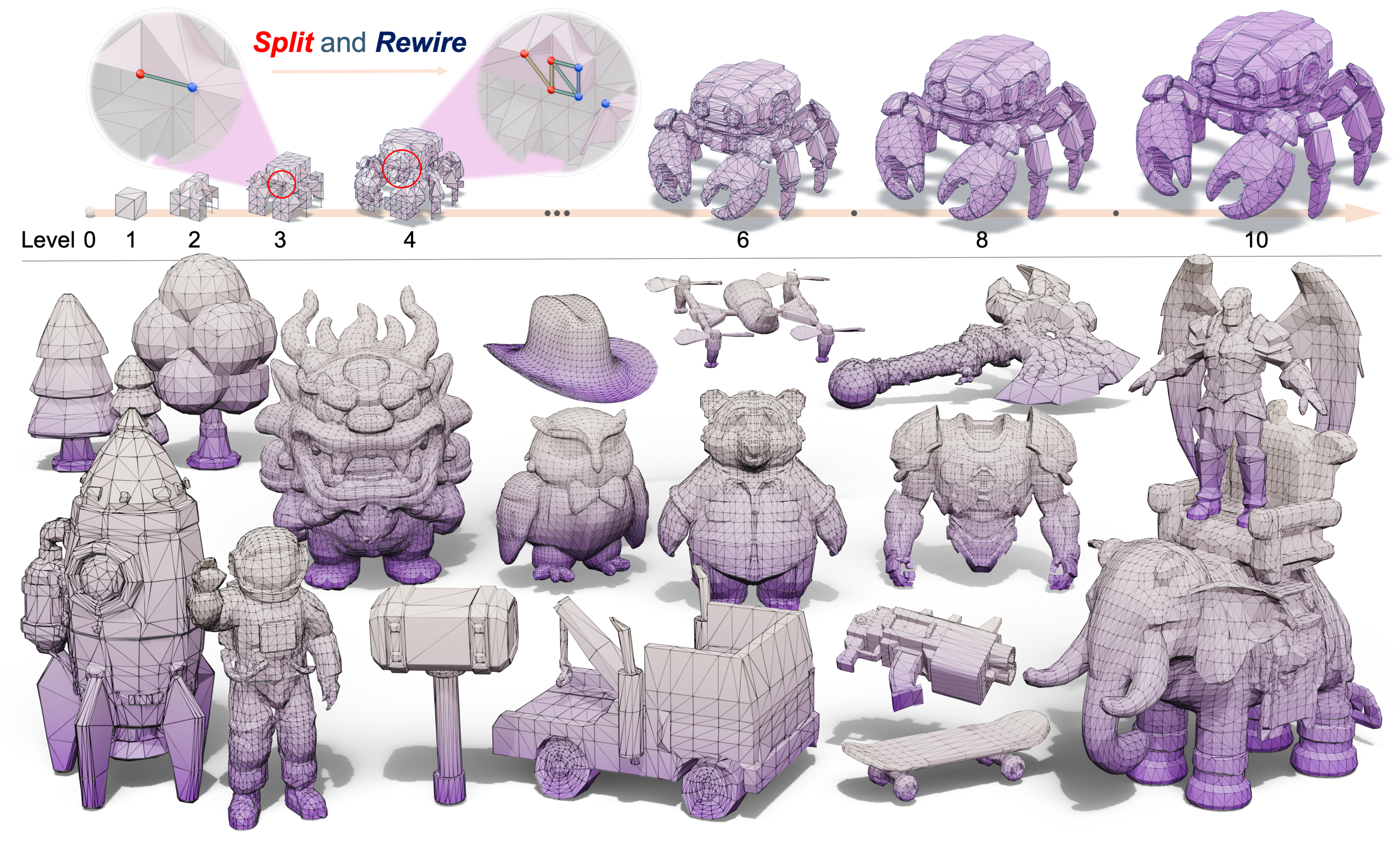}
\caption{\textbf{MeshOctave generates meshes via cascading resolution transitions.} Top: Starting from a single voxel, each step doubles vertex resolution by predicting child vertex instantiation (\textbf{split}) and reconnection (\textbf{rewire}) for all parent faces in parallel. 
Bottom: Meshes generated by MeshOctave.}
\label{fig:teaser}
\end{figure}

\begin{abstract}
Generating compact, artist-style meshes with explicit topology typically relies on autoregressive models which incur prohibitive sequential per-token costs, or continuous flow models that depend on heuristic connectivity decoders. Next-scale generation paradigms offer a compelling alternative by enabling parallel intra-scale token prediction and coarse-to-fine refinement from global structure to local topology; yet, existing methods derive hierarchical scales via progressive mesh simplification and invert them sequentially. This eliminates intra-scale parallelism and scales generation steps linearly with face count. In this paper, we propose \textbf{MeshOctave}, which instead defines scale through dyadic spatial grid resolutions, framing coarsening as a deterministic collapse that merges vertices sharing a voxel cell and inherits connectivity. Its inverse operation, \textbf{split-and-rewire}, determines which octant sub-vertices are instantiated for each coarse face and resolves local connectivity using discrete structural tokens. These per-face operations require no serialization, each scale transition is modeled as an unordered set that adds one bit of coordinate precision, naturally supporting dynamic-length meshes and adaptive resolution refinement. We construct a scale-conditioned masked-uniform discrete diffusion model to learn split-and-rewire operation from resolution collapse hierarchies. MeshOctave outperforms strong baselines in geometric fidelity and topological validity by a non-trivial margin, while supporting adaptive resolution refinement and extending naturally to mesh subdivision tasks.
\end{abstract}

\section{Introduction}
\label{sec:intro}
Meshes are the foundational geometric representation for computer graphics, animation, and simulation, where production assets require purposeful vertex layouts and well-structured edge connectivity. Recent 3D foundation models based on implicit fields~\citep{zhang20233dshape2vecset, zhang2024clay, zhao2023michelangelo, xiang2025structured, xiang2026native, zhao2025hunyuan3d, wu2026direct3d, li2025triposg, lai2026lattice, li2026sparc3d} achieve impressive shape fidelity, they decouple surface geometry from topology. Extracting explicit surfaces requires heuristic isosurface extraction~\citep{lorensen1998marching}, yielding over-tessellated meshes that necessitate costly manual retopology before downstream deployment. This disconnect has driven the shift toward native mesh generation with explicit topology.
%
Existing native generators broadly fall into two paradigms. Autoregressive (AR) models serialize meshes into sequential vertex and face tokens~\citep{nash2020polygen, siddiqui2024meshgpt, chen2025meshanything, chen2025meshanythingv2, hao2024meshtron, weng2025scaling, zhao2025deepmesh, lin2026meshripple}. While preserving discrete connectivity, their inference latency scales linearly with face count, compounding error propagation and imposing artificial ordering heuristics onto permutation-invariant graphs. Conversely, continuous diffusion and flow-matching models denoise mesh elements in parallel. They bypass discrete topology by converting topology to continuous surface fields~\citep{zhao2026lato, song2026mesh} or per-vertex embedding with spacetime supervision~\citep{li2026meshflow, wang2026nexus} or relaxed half-edges~\citep{shen2024spacemesh, xu2026meshy}. However, relying on post-hoc thresholding of continuous latents makes them fragile: minor predictive deviations precipitate missing or spurious faces. Furthermore, connectivity is resolved only at the terminal vertex set, even in coarse-to-fine architectures~\citep{wang2026nexus}. While discrete triangle-soup diffusion~\citep{alliegro2023polydiff, song2025tssr} avoids continuous relaxations, it reduces topology to mere coordinate coincidence. Crucially, all existing approaches resolve connectivity at a single monolithic scale: forcing models to predict macro-level geometry and micro-level topology simultaneously, inherently impairs their ability to balance global structural coherence with local geometric fidelity.

Next-scale prediction~\citep{tian2024visual} establishes a coarse-to-fine hierarchy that enables parallel intra-scale generation conditioned on coarse structural priors. While successfully extended to voxels~\citep{chen2025sar3d, medi20243d}, octrees~\citep{wei2025octgpt, feng2026grow3d}, point clouds~\citep{meng2026pointnsp}, and neural subdivision surfaces~\citep{guo2026subdivar}, adapting this paradigm to topology-preserving meshes remains an open challenge. It requires a coarsening operator that satisfies three key properties: (i) global parallelism, operating uniformly across the entire mesh so that scale transition is a single parallel pass; (ii) canonicity, the operation is an intrinsic, deterministic function of geometry rather than heuristic scheduling; and (iii) order-agnostic decodability, factorizes inverse refinement into locally independent decisions for concurrent decoding.
Recent coarse-to-fine models instead derive hierarchies from classical mesh simplification: VertexRegen~\citep{zhang2025vertexregen} inverts QEM-guided progressive edge collapses~\citep{garland1997surface, hoppe1999new}, while ARMesh~\citep{lei2026armesh} sequentially expands simplicial complexes from a base point. Both define scale as a sequence of individual local operations, violating all three criteria: the hierarchy reflects cost-metric tie-breaking rather than intrinsic shape, and refinements depend strictly on sequential decisions reinstating autoregressive latency,

To establish a canonical, parallelizable hierarchy for artist meshes, we define scale directly by spatial resolution. Halving the quantization grid resolution merges vertices falling within the same voxel, a grid-aligned form of vertex clustering~\citep{rossignac1993multi, luebke1997view} we term resolution collapse, with each merged vertex inheriting the combined connectivity of its constituents. Its inverse, \textbf{split-and-rewire}, factorizes into per-face decisions that determine: (i) which of the eight octant children of each vertex are instantiated at the finer resolution, and (ii) how these children connect internally and to the children of the face's other two vertices. Each transition adds one bit of coordinate precision per axis while simultaneously splitting merged vertices and rewiring their edges, reaching an $\mathrm{L}^3$ grid from a single root voxel in at most $\log_2 \mathrm{L}$ levels.
Leveraging this hierarchy, we introduce \textbf{MeshOctave}, a discrete diffusion model that generates meshes scale by scale (Fig.~\ref{fig:teaser}). For each coarse face, the split-and-rewire decisions are parameterized as a nine-element discrete structural tokens: three $8\times1$ binary occupancy vectors (one per parent vertex), three $8\times8$ intra-parent adjacency matrices, and three $8\times8$ inter-parent adjacency matrices. Because tokens are spawned directly from the faces generated at previous level, the sequence length dynamically adapts to geometric complexity across shapes and scales. Following TSSR~\citep{song2025tssr}, we corrupt these tokens with combined mask-based and uniform discrete noise~\citep{austin2021structured, sahoo2024simple, schiff2025simple} and denoise them using an hourglass transformer. Vertex occupancy tokens are decoded via categorical classification heads, whereas the six adjacency matrices are predicted row by row using a lightweight autoregressive head. Given a coarse mesh, the structural tokens for decoding next-scale mesh is then generated from a fully masked sequence, progressively refined through our propose-and-refine sampling loop until all structural tokens are confidently resolved. Extensive experiments demonstrate that MeshOctave outperforms strong autoregressive and flow-matching baselines by a substantial margin, demonstrating the feasibility and advantages of next-scale generation paradigm for native meshes.
Our primary contributions are then mainly: (1) \textbf{the formulation of split-and-rewire cascades} for canonical, globally parallel, and order-agnostic multi-scale mesh formulation; (2) \textbf{a scale adaptive discrete diffusion model} that supports next-scale native mesh generation and adaptive mesh resolution refinement.

\definecolor{MeshSpace}{HTML}{6699BE}
\definecolor{MeshModel}{HTML}{888888}
\definecolor{MeshSchedule}{HTML}{D5797C}
\providecommand{\meshbadge}[2]{%
  \tikz[baseline=(b.base)]{\node[circle,fill=#1,text=white,
    inner sep=0pt,outer sep=0pt,minimum size=3.4mm,
    font=\sffamily\bfseries\scriptsize] (b) {#2};}%
}
\providecommand{\meshspace}[1]{\meshbadge{MeshSpace}{#1}}
\providecommand{\meshmodel}[1]{\meshbadge{MeshModel}{#1}}
\providecommand{\meshschedule}[1]{\meshbadge{MeshSchedule}{#1}}
\providecommand{\meshcode}[3]{%
  \mbox{\meshspace{#1}\,\ensuremath{+}\,\meshmodel{#2}\,\meshschedule{#3}}%
}
\providecommand{\meshy}{\ensuremath{\checkmark}}
\providecommand{\meshn}{\ensuremath{\times}}

\begin{wraptable}{R}{0.52\textwidth}
\vspace{-\intextsep}
\vspace{-\abovecaptionskip}
\centering
\setlength{\intextsep}{8pt}

\renewcommand{\meshbadge}[2]{%
  \tikz[baseline=(b.base)]{%
    \node[
      circle,
      fill=#1,
      text=white,
      inner sep=0pt,
      outer sep=0pt,
      minimum size=2.5mm,
      font=\sffamily\bfseries\fontsize{5.8}{6}\selectfont
    ] (b) {#2};%
  }%
}

\renewcommand{\meshcode}[3]{%
  \mbox{%
    \meshspace{#1}\,\ensuremath{+}\,
    \meshmodel{#2}\,\meshschedule{#3}%
  }%
}

\caption[Difference with representative mesh generators.]{
{\textbf{Comparison with prior works.} MeshOctave jointly refines vertices and topology via multi-scale raw-space masked diffusion.}
\textbf{Notation:}
Rep. \meshspace{R}: raw;
\meshspace{L}: latent;
Method \meshmodel{A}: autoregressive;
\meshmodel{D}: diffusion/flow;
\meshmodel{M}: masked diffusion;
Arch. \meshschedule{N}: next scale;
\meshschedule{W}: whole mesh.
}
\label{tab:method_comparison}

\small
\setlength{\tabcolsep}{3pt}
\renewcommand{\arraystretch}{1.15}

\begin{adjustbox}{max width=\linewidth}
\begin{tabular}{@{}lccc@{}}
\toprule
Method & Orderless & Vertex & Topology \\
\midrule

MeshGPT & \meshn
& \multicolumn{2}{c}{\meshcode{L}{A}{W}} \\

BPT & \meshn
& \multicolumn{2}{c}{\meshcode{R}{A}{W}} \\

VertexRegen & \meshn
& \multicolumn{2}{c}{\meshcode{R}{A}{N}} \\

FastMesh & \meshn
& \meshcode{R}{A}{W} & -- \\

ARMesh & \meshn
& \multicolumn{2}{c}{\meshcode{R}{A}{N}} \\

TSSR & \meshn
& \multicolumn{2}{c}{\meshcode{R}{M}{W}} \\

MeshFlow & \meshy
& \multicolumn{2}{c}{\meshcode{L}{D}{W}} \\

LATO & \meshy
& \multicolumn{2}{c}{\meshcode{L}{D}{W}} \\

Nexus & \meshy
& \meshcode{R}{D}{N}
& \meshcode{L}{D}{W} \\

LATO.2 & \meshy
& \meshcode{L}{D}{W}
& \meshcode{L}{D}{W} \\

Meshy T2 & \meshy
& \multicolumn{2}{c}{\meshcode{L}{D}{W}} \\

\midrule
\textbf{MeshOctave} & \meshy
& \multicolumn{2}{c}{\meshcode{R}{M}{N}} \\

\bottomrule
\end{tabular}
\end{adjustbox}
\end{wraptable}


\section{Related Work}
\label{sec:related_work}

\paragraph{Autoregressive Mesh Generation.}
Autoregressive approaches formulate mesh synthesis as causal sequence modeling over quantized geometric and topological primitives. PolyGen~\citep{nash2020polygen} and MeshGPT~\citep{siddiqui2024meshgpt} established representative vertex-, face-, and latent-token formulations, followed by improvements in model capacity, conditioning~\citep{chen2024meshxl,chen2025meshanything,weng2024pivotmesh}. To avoid repeatedly encoding shared vertices, recent methods exploit vertex reuse~\citep{chen2025meshanythingv2}, connectivity-aware traversal~\citep{tang2025edgerunner,rossignac1999edgebreaker}, block and patch aggregation~\citep{weng2025scaling}, hierarchical BFS encoding~\citep{song2026meshsilksong}, or tree-structured sequencing~\citep{lionar2025treemeshgpt}. Other directions reduce decoding costs through hierarchical architectures~\citep{hao2024meshtron}, infer faces after vertex generation~\citep{kim2026fastmesh}, apply preference or reinforcement-based post-training~\citep{zhao2025deepmesh,liu2026mesh}, or extend autoregressive modeling to quadrilateral meshes~\citep{liu2026quadgpt}. MeshRipple~\citep{lin2026meshripple} further introduces frontier-aware BFS tokenization, expansive prediction, and NSCA for structured artist-mesh generation. These methods directly produce meshes with explicit connectivity, but retain sequential dependencies at the token, vertex, or local-update level, limiting generation parallelism as mesh complexity increases.

\paragraph{Diffusion-Based Mesh Generation.}
Diffusion and flow-matching models generate mesh representations through iterative updates that operate on multiple elements in parallel.
PolyDiff~\citep{alliegro2023polydiff} applies discrete diffusion to quantized triangle soups, while TSSR~\citep{song2025tssr} combines masked topology sculpting with uniform-noise shape refinement over mesh token sequences.
Continuous approaches learn latent representations of geometry and connectivity: LATO~\citep{zhao2026lato} uses structured voxel latents, MeshFlow~\citep{li2026meshflow} combines a mesh VAE with flow matching, and Meshy T2~\citep{xu2026meshy} jointly generates geometry and connectivity through vertex-set latents.
Geometry and topology can also be modeled in separate stages, as in LATO.2~\citep{long2026lato}, which generates vertices before sampling their conditional connectivity.
These methods avoid token-by-token decoding, but typically recover target-scale connectivity without conditioning it on an explicit coarser mesh.

\paragraph{Next-Scale Mesh Generation.}
VAR~\citep{tian2024visual} replaces next-token image prediction with next-scale prediction, generating all tokens at a scale conditioned on coarser scales.
Related 3D approaches explore multiscale latent representations and octree hierarchies, as in SAR3D~\citep{chen2025sar3d} and OctGPT~\citep{wei2025octgpt}.
For explicit meshes, progressive generation provides control over the level of detail: VertexRegen~\citep{zhang2025vertexregen} learns to reverse edge collapses through vertex splits, while ARMesh~\citep{lei2026armesh} grows simplicial complexes through local remeshing.
Both increase mesh detail through sequences of local operations.
Nexus~\citep{wang2026nexus} instead generates vertices through a spatial octree hierarchy and subsequently models topology conditioned on the generated vertices.
Our method defines mesh scales by dyadic spatial resolution and uses masked diffusion to predict split-and-rewire configurations in parallel within each scale.

\section{Vertex Split-and-Rewire Cascades}
\label{sec:method}
\begin{figure}[t]
\centering
\includegraphics[width=\textwidth]{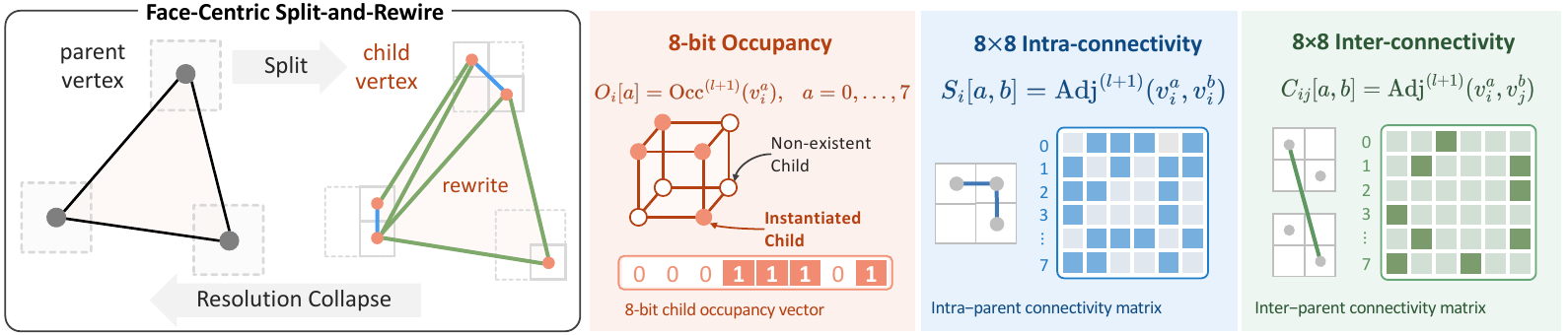}
\caption{\textbf{Face-centric split-and-rewire.}
Resolution collapse merges vertices fall in the same grid cell; its inverse, split-and-rewire specifies which of the eight child voxels are occupied and how the instantiated children are connected. For each parent face, 3 8-bit occupancy vectors encode vertex splitting, and 3 intra-parent and 3 inter-parent $8\times8$ connectivity matrices encode rewiring.}
\label{fig:tokenizer}
\end{figure}

Generating a whole scale in parallel requires a global coarsening operator whose inverse decomposes into local, order-free decisions. We derive from nested quantization grids.

\subsection{Resolution Collapse}
\label{sec:collapse}

Let a mesh at hierarchy level $l$ be defined as $\mathcal{M}^l = (\mathbf{V}^l, \mathbf{F}^l)$, where $\mathbf{V}^l \subset \mathbb{R}^3$ represents the set of vertices and $\mathbf{F}^l$ denotes the set of triangular faces. $\mathcal{E}(\mathcal{M}^l)$ denotes the induced edge set. Each face is defined as a triplet of vertices, $\mathbf{f}^l = \langle \mathbf{v}_0^l, \mathbf{v}_1^l, \mathbf{v}_2^l \rangle \in \mathbf{V}^l$. Vertex coordinates are quantized to a uniform grid of resolution $2^l$; thus, $l=10$ corresponds to a $1024^3$ grid. The resolution collapse operator $\mathbb{C}$ downsamples $\mathcal{M}^l$ to $\mathcal{M}^{l-1}$ by halving the grid resolution. During this process, each coarse voxel at level $l-1$ subsumes an octant of $2^3$ fine voxels at level $l$, denoted as mapping $\pi: \mathbf{V}^l \to \mathbf{V}^{l-1}$ that maps each fine vertex to its corresponding coarse level vertex. $\mathbf{F}^{l-1}$ are obtained by directly projecting each fine face via $\pi$, removing duplicates while preserving degenerated faces:
\begin{equation}
  \mathbf{V}^{l-1} \leftarrow \pi(\mathbf{V}^l), \,\,\, \mathbf{F}^{l-1} \leftarrow \bigl\{ \bigl \langle \pi(\mathbf{v}_0^l), \pi(\mathbf{v}_1^l), \pi(\mathbf{v}_2^l) \bigr \rangle \;\big|\;  \langle \mathbf{v}_0^l, \mathbf{v}_1^l, \mathbf{v}_2^l \rangle \in \mathbf{F}^l \bigr\}.
  \label{eq:face_collapse}
\end{equation}
We use operator $\mathbb{C}$ to denote Eqn.~\ref{eq:face_collapse}, which operates independently and uniformly across all grid voxels in a single pass, the resulting multiscale hierarchy,
\begin{equation}
  \mathcal{M}^{\mathrm{L}}
  \xrightarrow{\ \mathbb{C}\ }
  \mathcal{M}^{\mathrm{L}-1}
  \xrightarrow{\ \mathbb{C}\ }
  \cdots
  \xrightarrow{\ \mathbb{C}\ }
  \mathcal{M}^{0},
  \label{eq:hierarchy}
\end{equation}
is canonical and determined by the initial mesh and the spatial grid.

\subsection{Face-centric Split-and-Rewire Cascades} 
Next-scale mesh generation then is the inversion process of resolution collapse, hence it is essential to define the inverse of $\mathbb{C}$. This inverse process $\mathbb{S}$ doubles the spatial resolution and recover the information discarded: which children of each parent vertex exist, and how they connect.

\noindent\textbf{Vertex Split.}
With grid resolution doubled, a given parent vertex $\mathbf{v}^{l-1}$ is divided into $2^3$ sub-voxels (potentially contain child-vertex) in level $l$, yielding an octant $\mathcal{O}(\mathbf{v}^{l-1})$. 
We directly encode the instantiation of child-vertex into an $8$-bit binary vector $\mathbf{O}(\mathbf{v}^{l-1}) \in \{0, 1\}^8$ with $\mathbf{O}(\mathbf{v}^{l-1})=\mathbf{1}[\mathcal{O}(\mathbf{v}^{l-1}) \in \mathbf{V}^l]$, i.e., depending on whether the sub-voxels presents as vertices in $\mathbf{V}^l$. Thus, a single vector specifies the complete subdivision pattern for a given parent vertex.

\noindent\textbf{Rewire.}
We must also recover the connectivity of the split vertices. Since each parent vertex yields up to eight children, the connectivity between the children of any two parent vertices can be modeled using an $8\times8$ binary matrix $\mathbf{M}$. 
For a given parent face $\mathbf{f}^{l-1} = \langle \mathbf{v}_0^{l-1}, \mathbf{v}_1^{l-1}, \mathbf{v}_2^{l-1} \rangle$, we define the connectivity matrix $\mathbf{M}_{ij}$ between the children of parent vertices $\mathbf{v}_i^{l-1}$ and $\mathbf{v}_j^{l-1}$ as:
\begin{equation}
    \mathbf{M}_{ij}(a,b)
    =
    \mathbf{1}
    \left[
        \bigl < \mathcal{O}_a(\mathbf{v}_i^{l-1}), \mathcal{O}_b(\mathbf{v}_j^{l-1}) \bigr >
        \in \mathcal{E}(\mathcal{M}^l)
    \right],
    \,\,
    0 \leq i \leq j \leq 2,
    \label{eq:connectivity}
\end{equation}
where $a, b \in \{0, \dots, 7\}$ index the sub-voxel (child vertex).
When $i=j$, the matrix records connections among the children of the same parent vertex, which we denote as the intra-parent connectivity matrix $\mathbf{S}_{i}$. When $i < j$, the matrix records connections between the children of two distinct parent vertices, denoted as the inter-parent connectivity matrix $\mathbf{C}_{ij}$. Consequently, a parent face contains three intra-parent and three inter-parent connectivity matrices.

Together, the inversion of collapsing any parent face $\mathbf{f}^{l-1}$ is fully represented by nine-element structural token $\mathbf{z}$ (three vertex occupancy vectors and six connectivity matrices).
\begin{equation}
\mathcal{M}^{l-1}  \xrightarrow{\ \mathbb{S}(\mathbf{z}) \ }  \mathcal{M}^{l}; \quad
\mathbf{z}_{\mathbf{f}^{l-1}}
=
\bigl(
\mathbf{O}_0, \mathbf{O}_1, \mathbf{O}_2;\,
\mathbf{S}_0,
\mathbf{S}_1,
\mathbf{S}_2;\,
\mathbf{C}_{01},
\mathbf{C}_{02},
\mathbf{C}_{12}
\bigr).
\label{eq:ninetokens}
\end{equation}  
During each resolution collapse, we record $\mathbf{z}$ for every face. And fine mesh $\mathcal{M}^\mathrm{L}$ can by fully recovered from a single voxel at level 0 through face-wise split-and-rewire operation $\mathbb{S}$.

\section{Scale Adaptive Discrete Diffusion}
\label{sec:model}
\begin{figure}[t]
\centering
\includegraphics[width=\textwidth]{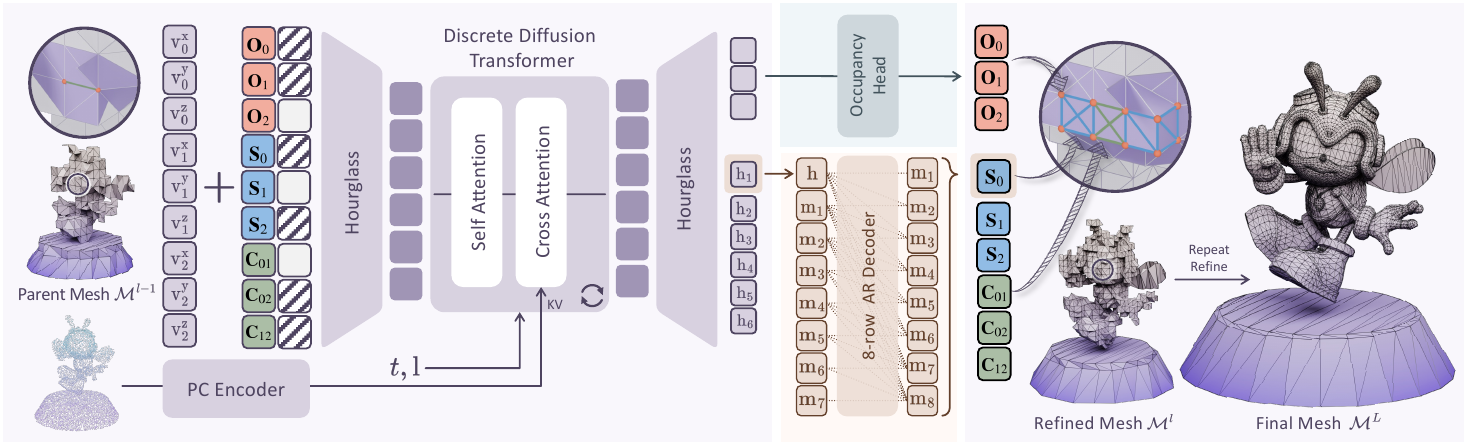}
\caption{\textbf{Scale-conditioned discrete diffusion model.}
The generator refines a mesh from scale $l$ to $l+1$, the split-and-rewire structural tokens $\mathbf{z}$ is concatenated with vertex coordinates. The elements of $\mathbf{z}$ are masked according to diffusion timestep $t$. An hourglass Transformer processes these tokens, modulated by timestep $t$ and scale $l$ via adaLN. Finally, categorical heads predict child-vertex occupancy while an autoregressive MLP head decodes row-factorized adjacency matrices, assembling the refined mesh at scale $l+1$.}
\label{fig:pipeline}
\end{figure}

We adopt the split-and-rewire cascade (Eqn.~\ref{eq:ninetokens}) for next-scale mesh generation: given a coarse mesh $\mathcal{M}^{l-1}$, the generator predicts a structural token $\mathbf{z}$ for each face $\mathbf{f}^{l-1}$. Since $\mathbf{z}$ contains only binary vectors and matrices, we formulate its prediction as a discrete diffusion process, in which the structural token of each face is progressively revealed from masked tokens.

To input the coarse model $\mathcal{M}^{l-1}$, we fuse $\mathbf{z}$ with the vertex coordinates of its corresponding face in $\mathcal{M}^{l-1}$. As both contain nine elements per face, they are added directly after embedding: coordinate slot $k$ of face $f$ add with structural token slot $z_{f,k}$: $\mathbf{x}_{f,k} = \textrm{Embed}(z_{f,k}) + \text{Embed}(\mathbf{f}[k])$, we omit $l-1$ for clearity.
where $\text{Embed}(\cdot)$ is embedding, and $\mathbf{x}_{f,k}$ is the resulting slot-wise input to the network. An hourglass encoder~\cite{hao2024meshtron} then compresses $\mathbf{x}_{f,k}$ into vertex-level and subsequently face-level tokens. A transformer backbone processes the face-level tokens, and an hourglass decoder expands its output back to nine tokens per face. Finally, an occupancy decoder $\mathcal{D}_o$ and a connectivity decoder $\mathcal{D}_c$ map the hidden states to $\mathbf{z}$. $\mathcal{D}_o$ is a $256$-way classifier whose output index maps to the $8 \times 1$ binary occupancy $\mathbf{O}$. $\mathcal{D}_c$ is a small MLP autoregressively predicts the eight rows $\mathbf{m}_{r}, r=1...8$ of the connectivity matrix $\mathbf{M}$ (Eq.~\ref{eq:connectivity}), where each step conditions on output hidden $\mathbf{h}_m$ corresponding to $\mathbf{S}$ and $\mathbf{C}$, and the previously decoded rows:
\begin{equation}
    \mathcal{D}_c: \, p_c (\mathbf{M} \mid \mathbf{h}_m)
    = \prod_{r=1}^{8}
    p_c(\mathbf{m}_{r}\mid\mathbf{m}_{<r},\mathbf{h}_m).
    \label{eq:matrixhead}
\end{equation}
In the diffusion process, each connectivity token $\mathbf{M}$ is nonetheless treated as a single slot, with all eight rows masked or revealed together.

\subsection{Masked-Uniform Discrete Diffusion.}
\label{sec:diffusion-process}
We adopt a discrete-state diffusion framework~\citep{austin2021structured}, in which each slot in $\mathbf{z}$ is corrupted independently with probability $r(t)=\cos(\pi t/2)$, $t\sim\mathcal{U}(0,1)$, so that $t=0$ corresponds to full corruption and $t=1$ to clean tokens. We adopt the standard $\mathbf{z}_0$-parameterization and train the network to recover the clean tokens directly from the corrupted sequence $\tilde{\mathbf{z}}$:
\begin{equation}
    \mathcal{L}(\theta) = \mathbb{E}_{t,\,l,\,\tilde{\mathbf{z}}}\Big[\sum_{(f,k)\in\Omega}
    -\log p_\theta\big(z^\star_{f,k}\mid \tilde{\mathbf{z}}, \mathbf{F}^{l-1}, t, l, \mathbf{c}\big)\Big],
    \label{eq:simple-loss}
\end{equation}
where $z^\star_{f,k}$ denotes the ground-truth value of the $k$-th slot of $\mathbf{z}$ corresponding to face $f$, and $p_\theta$ is parameterized per slot by either the categorical occupancy head or the autoregressive connectivity head
(Eq.~\ref{eq:matrixhead}). Following standard practice in mesh generation~\citep{chen2025meshanything, wang2026nexus}, we condition the generator on point-cloud $\mathbf{c}$, and $\Omega$ denotes the set of slots on which the loss is computed.

\paragraph{Mask-Uniform Noise Strategy.}
\label{sec:mask-uniform}

In standard masked diffusion, unmasked tokens are permanently fixed, incurring early prediction errors to irreversibly propagate into the final mesh. Inspired by TSSR~\citep{song2025tssr}, we adopt a mask-uniform noising strategy: a mask pathway that predicts structure from partial tokens, and a uniform pathway that refines the fully populated sequence. Both share a single backbone and are conditioned via a mode flag $\mathtt{[MODE]}$ appended to the point-cloud feature $\mathbf{c}$.

The mask pathway replaces corrupted slots ${f,k}\in \mathcal{B}$ with $\mathtt{[MASK]}$ and is supervised only on those slots, then we set $\Omega=\mathcal{B}$ and compute the loss $\mathcal{L}_{\text{mask}}$ from Eq.~\ref{eq:simple-loss}. The uniform pathway instead replaces corrupted slots with values drawn uniformly from the slot's state space and is supervised on all slots, i.e., $\Omega=\{1,\dots,9n\}$, and we compute $\mathcal{L}_{\text{unif}}$ from Eq.~\ref{eq:simple-loss}. Since the corrupted positions are not indicated, the model must learn to detect and correct any errors. Both losses use a multi-class focal loss~\citep{Linfocal2017} instead of cross-entropy to mitigate class imbalance by down-weighting well-classified slots, details see \ref{sec:appendix-focal}. The uniform pathway additionally trains a confidence head $\mathcal{C}_\phi$ to predict whether its own prediction at each slot is correct, with label
$y_{f,k}=\mathbf{1}\big[\arg\max p_\theta(\cdot\mid\mathbf{h}_{f,k}) = z^\star_{f,k}\big]$.
The head operates on $\mathrm{sg}[\mathbf{h}_{f,k}]$, so that
$\mathcal{L}_{\text{conf}}$ does not update the backbone:
\begin{equation}
    \mathcal{L}_{\text{conf}} = \frac{1}{9n}\sum_{f,k}
    \kappa_{f,k}\,\mathrm{BCE}\big(\mathcal{C}_\phi(\mathrm{sg}[\mathbf{h}_{f,k}]),\,y_{f,k}\big),
    \label{eq:lconf}
\end{equation}
where $n$ is the number of faces, and $\kappa_{f,k}=\kappa>1$ for the minority incorrect class ($y_{f,k}=0$) and $1$ otherwise. To expose the uniform pathway to realistic errors, a fraction $\rho$ of its training steps replaces uniform noise with the model's own greedy predictions from a gradient-free mask-pathway forward pass. At each step, we draw $s\sim\mathrm{Bernoulli}(p)$ with $p=0.5$ and minimize
\begin{equation}
    \mathcal{L}_\text{total} =
    s\,\mathcal{L}_{\text{mask}} + (1-s)\big(\mathcal{L}_{\text{unif}} + \lambda\,\mathcal{L}_{\text{conf}}\big),
    \label{eq:ltotal}
\end{equation}
where $\lambda$ balances the confidence loss.

\paragraph{Vertex-anchored 3D RoPE.}
\label{sec:rope}
Unlike TSSR’s 1D indexing, we represent meshes via orderless per-face token blocks to preserve block permutation equivariance. We inject structure via 64d RoPE, allocating 30 frequency pairs to 3D voxel coordinates and 2 to intra-face roles $\tau$ (occupancy, intra-/inter-parent connectivity). Tokens anchor to vertex pairs $(v_a, v_b)$ or $(v_a, v_a)$ for vertices and connectivity resprectively (Table~\ref{tab:anchors}). Sharing rotation angles across half their geometric frequencies makes vertex-edge incidence an explicit phase alignment. Finally, lexicographical ($z$-$y$-$x$) parent sorting canonicalizes shared edges, and reduced coarse tokens remain anchored to all three parent vertices.

\paragraph{Inference.}
During inference, the two pathways alternate within an iterative propose-and-correct loop. Initialized from a fully masked sequence, the mask pathway first generates candidate structures for all uncommitted regions of the mesh. The uniform pathway then evaluates the entire sequence, simultaneously refining token representations and assigning confidence scores via $\mathcal{C}_\phi$. Rather than permanently freezing unmasked tokens with a hard threshold, slots are stochastically re-masked ($\mathtt{[MASK]}$) based on their confidence scores and an annealed decoding schedule. Tokens with lower confidence are re-masked with higher probability, enabling the network to re-evaluate early errors while progressively committing to stable structures in later iterations. 
This cycle of proposal and refinement continues until all sampling steps are completed.

\subsection{Implementation.} 

We instantiate our discrete diffusion backbone as a symmetric hourglass Transformer with $9/3/1$ tokens per face, totaling $24$ blocks across stage depths $[2,4,6,6,4,2]$, width $1024$, $16$ heads, SwiGLU FFNs of width $2816$, and RMSNorm, for $624$M parameters. We adopt vertex-anchored 3D RoPE and variable-length FlashAttention-2 throughout. Conditioning combines adaLN-Zero modulation, the sum of timestep $t$ and scale level $l$ embeddings, with cross-attention over point-cloud features $\mathbf{c}$.
Condition $\mathbf{c}$ is derived via an 8-layer Michelangelo-style encoder of width $768$ with $12$ heads and $1024$ latents, totaling $193$M parameters and fine-tuned jointly with the backbone. Inputs consist of 40960 point-normal pairs subsampled from 50k surface samples, augmented with random scaling and Gaussian jitter of $\sigma=0.01$ applied with probability $0.5$. We train on $350$K meshes filtered from the Objaverse and Toys4K datasets, up to $15$k faces and $135$k tokens each at the finest $9$-token/face resolution, for 7 days on 8$\times$NVIDIA H800 80GB GPUs using bfloat16 mixed precision, and gradient checkpointing.

\vspace{-6pt}
\section{Experiments}
\label{sec:experiments}
\noindent\textbf{Evaluation Protocol.}
To ensure a fair comparison of \method{} with existing methods without bias, we select 300 meshes from ObjaverseXL and Toys4K, generate 200 meshes using large-scale 3D generative models~\citep{xiang2025structured,hunyuan3d2025hunyuan3d,wu2026direct3d} as our test set. We uniformly sample 4,096 surface points from each generated mesh and its reference mesh. Geometric fidelity is measured using CD-L1, CD-L2, and Hausdorff distance. We additionally report absolute normal consistency to evaluate surface orientation and local structure.

\noindent\textbf{Baselines.}
We compare with representative mesh generation methods across three paradigms. The autoregressive baselines include MeshAnythingV2~\cite{chen2025meshanythingv2}, MeshSilksong~\cite{song2026meshsilksong}, BPT~\cite{weng2025scaling}, DeepMesh~\cite{zhao2025deepmesh}, FastMesh~\cite{kim2026fastmesh}, and MeshRipple~\cite{lin2026meshripple}. The next-scale baselines include VertexRegen~\cite{zhang2025vertexregen} and ARMesh~\cite{lei2026armesh}. The flow-matching baselines include MeshFlow~\cite{li2026meshflow}, LATO~\cite{zhao2026lato}, and LATO.2~\cite{long2026lato}.

\begin{table*}[t]
\centering
\caption{\textbf{Quantitative comparison.}
\emph{Type} denotes the generative paradigm: autoregressive generation (AR), continuous flow matching (Flow), or discrete 
diffusion (DD). \Ntag{} marks methods that generate meshes over a coarse-to-fine resolution hierarchy and expose intermediate scales. Best results are in \textbf{bold}, and second-best results are
\underline{underlined}.}
\label{tab:quant}

\footnotesize
\setlength{\tabcolsep}{1.5pt}
\renewcommand{\arraystretch}{1.15}
\renewcommand{\methodhead}[1]{%
    \begingroup
    \renewcommand{\arraystretch}{0.9}%
    \begin{tabular}[c]{@{}c@{}}
        #1
    \end{tabular}%
    \endgroup
}
\begin{tabular}{@{}l*{12}{c}@{}}

\toprule
\methodhead{Metric}
& \methodhead{Mesh\\Any.V2}
& \methodhead{Vertex\\Regen}
& \methodhead{AR\\Mesh}
& \methodhead{Fast\\Mesh}
& \methodhead{Mesh\\Silk.}
& \methodhead{BPT}
& \methodhead{Deep\\Mesh}
& \methodhead{Mesh\\Ripple}
& \methodhead{Mesh\\Flow}
& \methodhead{LATO}
& \methodhead{LATO.2}
& \methodhead{MESH\\OCTAVE} \\
\midrule

Type
& AR
& AR+\Ntag
& AR+\Ntag
& AR
& AR
& AR
& AR
& AR
& Flow
& Flow
& Flow
& DD+\Ntag \\

\midrule

CD-L2 $\downarrow$
& 0.1103
& 0.0874
& 0.0804
& 0.0647
& 0.0622
& 0.0610
& 0.0505
& 0.0468
& 0.0452
& 0.0430
& \underline{0.0406}
& \textbf{0.0392} \\

CD-L1 $\downarrow$
& 0.1539
& 0.1238
& 0.1146
& 0.0925
& 0.0895
& 0.0873
& 0.0729
& 0.0691
& 0.0672
& 0.0621
& \underline{0.0595}
& \textbf{0.0581} \\

HD $\downarrow$
& 0.2317
& 0.1875
& 0.1569
& 0.1151
& 0.1440
& 0.1111
& 0.0957
& 0.0948
& 0.0765
& 0.0741
& \underline{0.0665}
& \textbf{0.0645} \\

$|\mathrm{NC}|$ $\uparrow$
& 0.6807
& 0.6655
& 0.6058
& 0.7413
& 0.7728
& 0.7976
& 0.8187
& 0.8141
& 0.8237
& 0.8273
& \underline{0.8333}
& \textbf{0.8478} \\

\bottomrule
\end{tabular}
\end{table*}
\vspace{-4pt}
\subsection{Quantitative Analysis.}
\label{sec:quant}
\textbf{Quality} \method{} outperforms all autoregressive, next-scale, and flow-matching baselines across every metric (Table~\ref{tab:quant}). Although flow-matching methods generally surpass autoregressive models by avoiding surface fragmentation and ensuring completeness, \method{} outperforms the strongest flow models by a non-trivial margin.

\textbf{Inference Latency.} \method{} takes $\sim$8\,s per scale (20 diffusion steps) and $\sim$90\,s across 10 scales, nearly independent from face count. In contrast, baselines such as ARMesh and VertexRegen require $\sim$15 minutes for 5k faces, a $\sim$10$\times$ speedup that highlights the efficiency of our parallel framework (Appendix~\ref{app:speed}).
%

\vspace{-4pt}
\subsection{Qualitative Analysis.}

\begin{figure}[t]
\centering
\includegraphics[width=\textwidth]{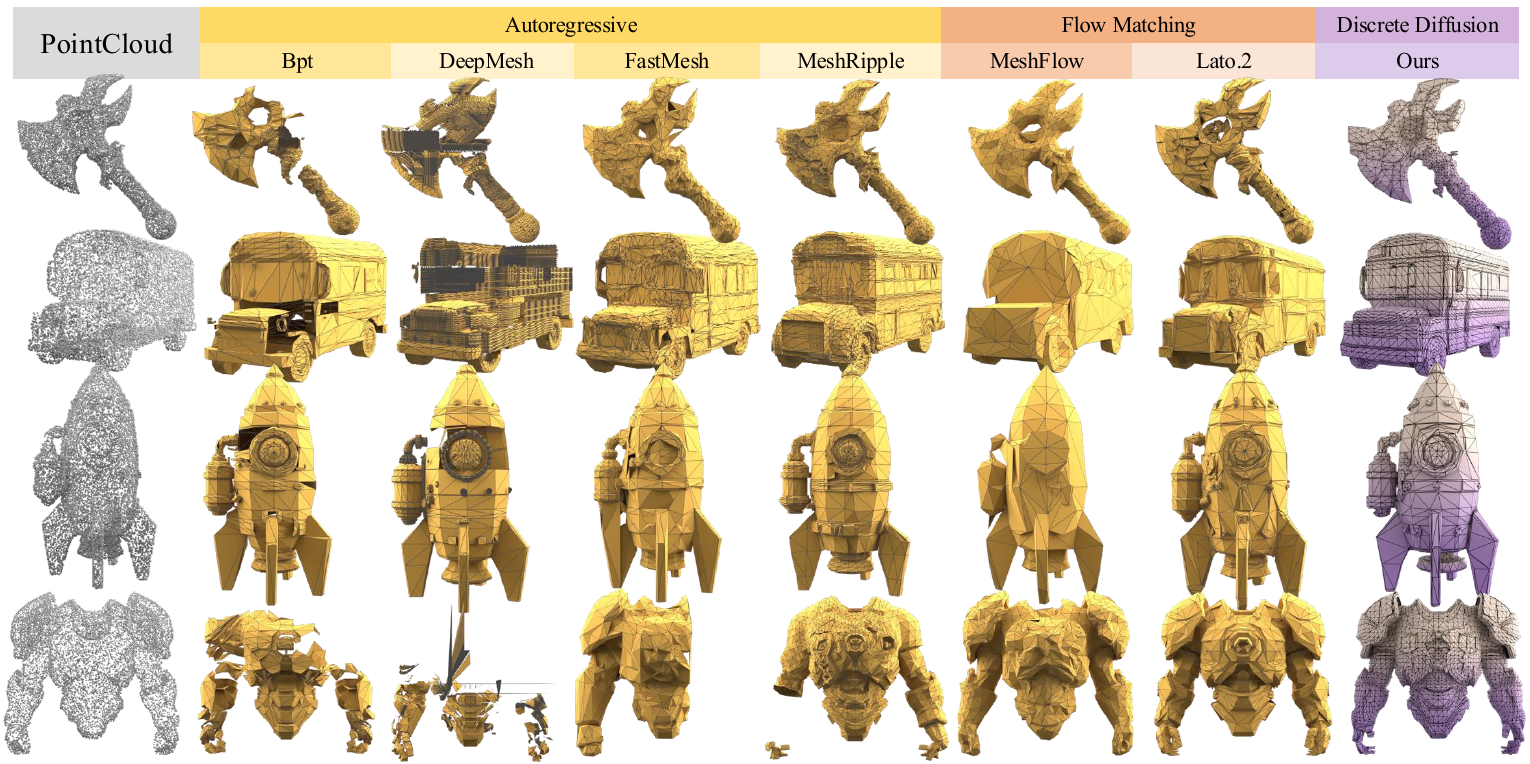}
\caption{\textbf{Qualitative comparison.} Each row shows one input point cloud and the final meshes generated by autoregressive, flow-matching, and diffusion-based methods.}
\label{fig:qual}
\end{figure}

\begin{wrapfigure}[15]{l}{0.45\textwidth}
\centering
\includegraphics[width=0.43\textwidth]{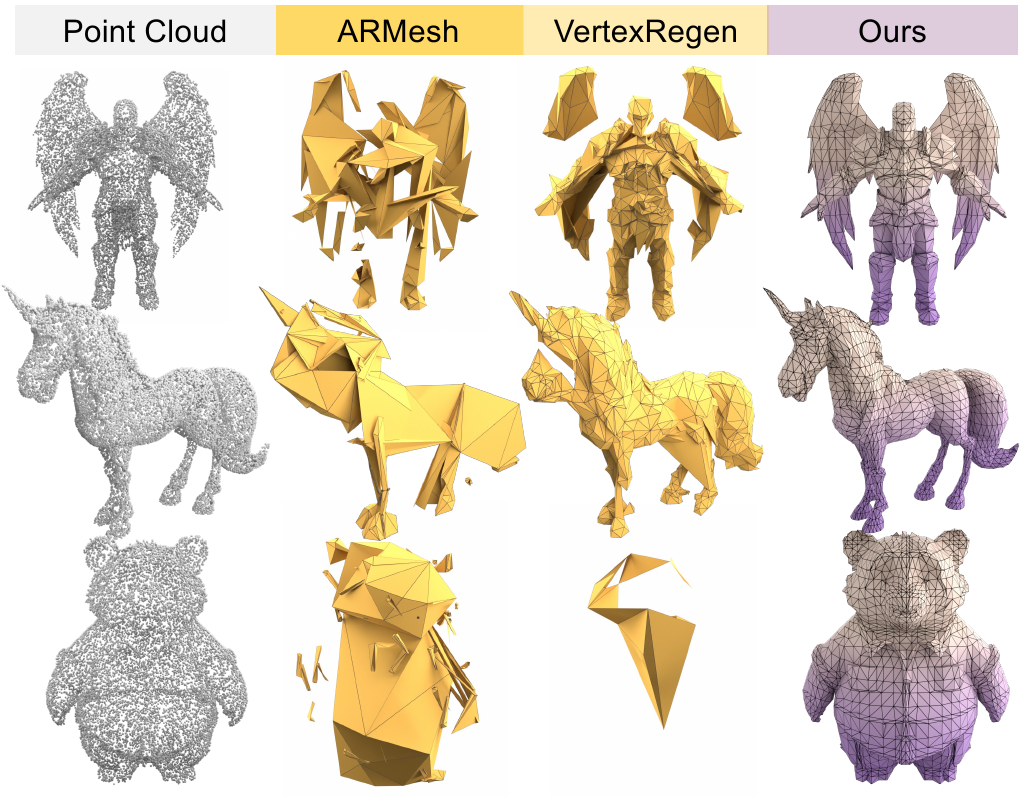}
\caption{\textbf{Comparison with level-of-detail baselines.}
}
\label{fig:qual_lod}
\end{wrapfigure}

\vspace{-\intextsep}
Figure~\ref{fig:qual} presents qualitative comparisons. While autoregressive baselines frequently show broken surfaces, flow-matching approaches tend to oversmooth fine geometric details. In contrast, \method{} faithfully preserves intricate shape features and delicate parts.

\begin{figure}[t]
\centering
\includegraphics[width=\textwidth]{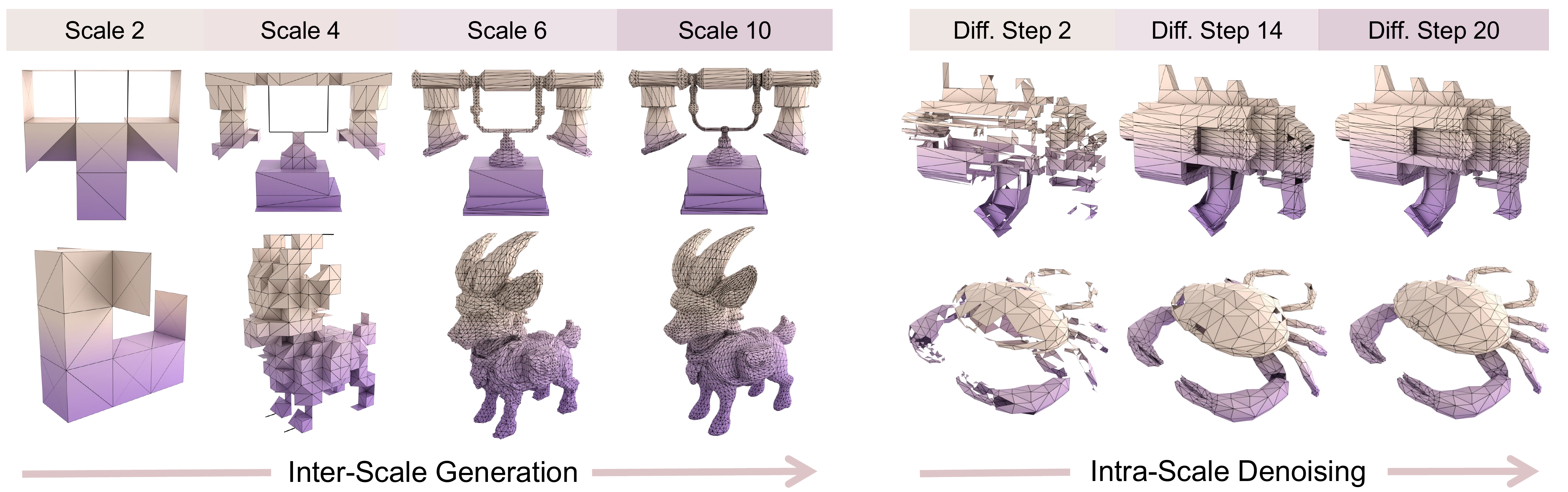}
\caption{\textbf{Generation process visualization.} Meshes are progressively generated across scales, while
predictions within a scale are iteratively refined through masked and uniform denoising.}
\label{fig:inter-intra-generation}
\end{figure}

Figure~\ref{fig:inter-intra-generation} visualizes both inter-scale generation and intra-scale denoising process.

Fig.~\ref{fig:qual_lod} compares \method{} with VertexRegen and ARMesh, the coarse-to-fine baselines closest to our paradigm. Lacking public weights, we retrained both from scratch on 200k meshes ($<5\text{k}$ faces) for five days under their official configurations. Relying on sequential, single-element refinements, both baselines suffer from compounding errors that cause severe geometric distortions. Conversely, \method{} resolves split-and-rewire decisions for entire resolution levels in parallel, producing substantially more coherent and watertight meshes.

\subsection{Ablation Study.}
\label{sec:ablation}
\begin{figure}[h]
\begin{minipage}[t]{0.40\textwidth}
\vspace{0pt}
\captionof{table}{\textbf{Ablation study.} One component replaced per row; best in \textbf{bold}.}
\label{tab:ablation}
\centering
\footnotesize
\setlength{\tabcolsep}{2.5pt}
\begin{tabular}{@{}lccc@{}}
\toprule
Variant & CD-L2 $\downarrow$ & HD $\downarrow$ & $|\mathrm{NC}|$ $\uparrow$ \\
\midrule
Full            & \textbf{0.0430} & \textbf{0.0739} & \textbf{0.8379} \\
\midrule
Continuous    & 0.0512 & 0.0983 & 0.6685 \\
w/o Uniform   & 0.0453 & 0.0884 & 0.7955 \\
w/o AR head   & 0.0496 & 0.1005 & 0.7907 \\
w/o 3D RoPE   & 0.0469 & 0.0915 & 0.7911 \\
\bottomrule
\end{tabular}
\end{minipage}
\hfill
\begin{minipage}[t]{0.60\textwidth}
\vspace{5pt}
\centering
\includegraphics[width=\linewidth]{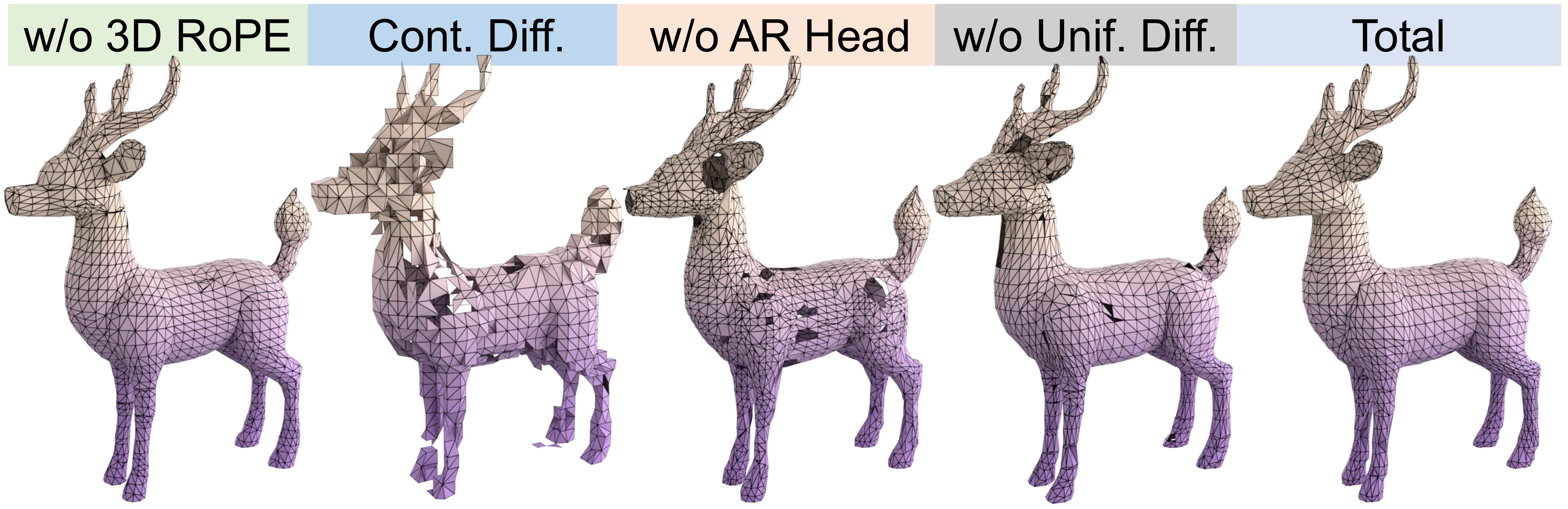}
\captionof{figure}{\textbf{Ablation, qualitative.} The same shape produced by each variant,
labelled above each panel.}
\label{fig:ablation}
\end{minipage}
\end{figure}

We ablate our core design choices in Table~\ref{tab:ablation} by evaluating four variants: replacing discrete with continuous diffusion (\text{Continuous}), omitting the uniform pathway (\text{w/o Uniform}), predicting adjacency rows independently without the AR head (\text{w/o AR Head}), and substituting 3D RoPE with 1D RoPE (\text{1D RoPE}). Continuous diffusion performs worst, as relaxing discrete constraints causes severe topological collapse. Mask-only diffusion lacks a refinement mechanism to correct early commitments. Predicting connectivity rows independently results in spurious edge connections. 1D RoPE discards explicit 3D coordinates, substantially degrading the model's awareness of local geometric proximity.


\subsection{Application.}
\begin{figure}[h]
\centering
\includegraphics[width=\textwidth]{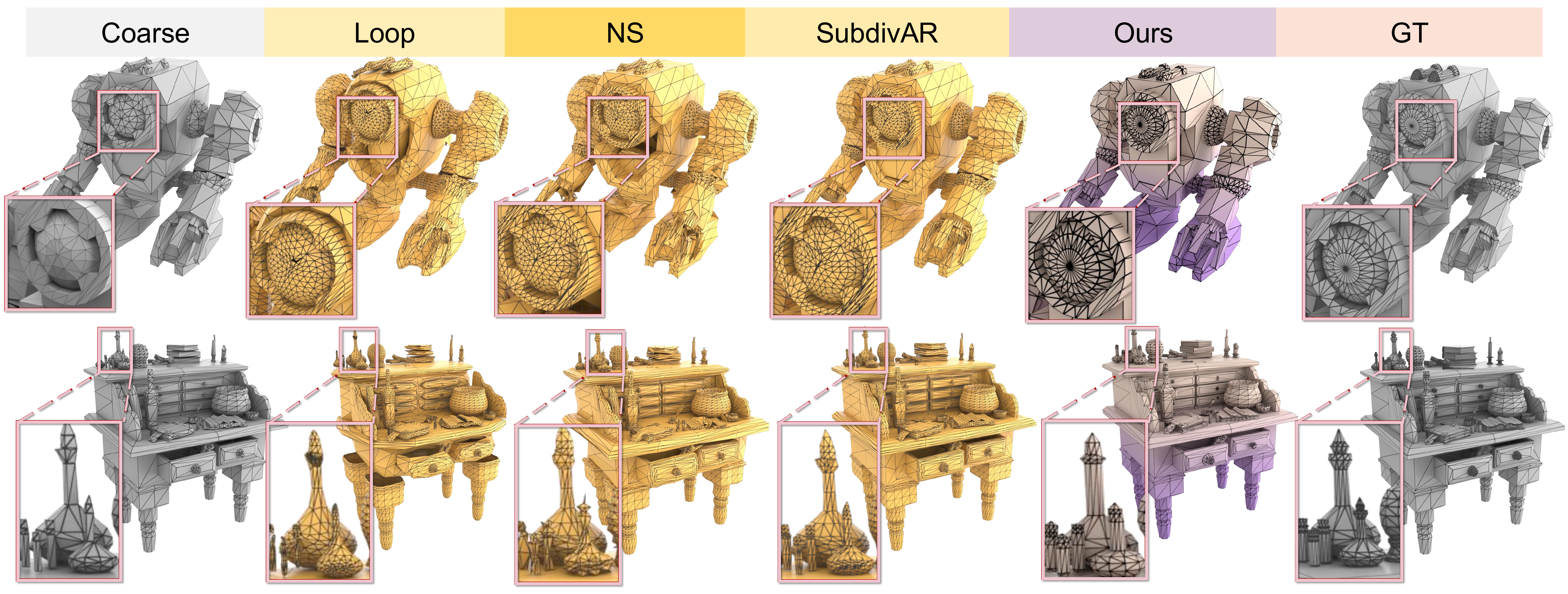}
\caption{\textbf{Application in mesh subdivision.} \method{} adaptively refines geometry and connectivity given an existing coarse mesh, producing more detailed and artist-like subdivision results.}

\label{fig:application}
\end{figure}

\textbf{Mesh Subdivision} \method{} naturally performs mesh subdivision without retraining by conditioning on points sampled directly from the coarse input (Fig.~\ref{fig:application}). Unlike baselines~\citep{loop1987smooth,liu2020neural,guo2026subdivar} that enforce uniform, fixed-topology subdivision, \method{} adaptively predicts child occupancy and connectivity—modifying topology and concentrating faces on complex regions for superior geometric fidelity.

\section{Conclusion}
We present \method{}, a next-scale generative framework for native 3D meshes centered on a globally parallel, order-agnostic split-and-rewire cascade. Driven by a mask-uniform discrete diffusion model across resolution scales, \method{} predicts intra-scale structural tokens in parallel, bypassing the sequential bottlenecks of prior multiscale methods. While generating sequentially across scales yields higher latency than single-scale flow matching and risks propagating coarse topological errors without backtracking, future work will investigate diffusion distillation for faster inference and bidirectional transitions to dynamically revise coarse topology.

\section*{AI use statement}
In preparing this manuscript, the authors employed ChatGPT exclusively as an editorial aid to polish English expression, correct grammatical errors, and enhance clarity. It was not involved in data collection or processing, conceptual or theoretical development, hypothesis formulation, mathematical reasoning, methodology or experimental design, method implementation, translation, qualitative analysis, or the interpretation of experimental findings. All suggested revisions were carefully examined and, when necessary, modified by the authors to maintain technical accuracy and the original intent. The authors assume full responsibility for the integrity and final content of this manuscript.

\bibliography{iclr2027_conference}
\bibliographystyle{iclr2027_conference}
\newpage
\appendix
\section{Appendix}
\subsection{Details of the Split-and-Rewire Tokenizer}
\label{sec:appendix-tokenizer}

\paragraph{Octant-absolute indexing.} Rows and columns of $\mathbf{S}_i$ and
$\mathbf{C}_{ij}$ are indexed by the child's fixed octant id $a\in\{0,\dots,7\}$
rather than by a compacted position among the children actually present. This
makes every matrix a full, lossless $8\times 8$ byte regardless of how many
children a parent vertex has, and removes any need to merge, reorder, or budget
children when a parent has close to the maximum of eight -- a parent with all
eight children behaves identically to one with a single child, just with a
denser matrix.

\paragraph{Occupancy disambiguates connectivity.} $\mathbf{S}_i$ and
$\mathbf{C}_{ij}$ are only meaningful for octants that $\mathbf{O}$ marks
active: at decode time every row and column of $\mathbf{S}_i$ is masked by
$\mathbf{O}_i$, and every row of $\mathbf{C}_{ij}$ by $\mathbf{O}_i\otimes
\mathbf{O}_j$, before any edge is read out. This keeps occupancy and
connectivity strictly separable in the representation: the generator can
predict a connectivity byte for an octant slot without that byte being
mistaken for evidence that the slot is occupied.

\paragraph{Degenerate faces.} A predicted edge does not always close into a
triangle at a given level -- a thin feature that occupies a single voxel
presents as a dangling edge or an isolated active octant rather than three
mutually connected ones. Rather than dropping such connectivity, we decode it
as a degenerate, zero-area face: an unclosed edge becomes $(A,B,B)$ and an
isolated active octant becomes $(A,A,A)$, using the same
$\mathbf{O}/\mathbf{S}_i/\mathbf{C}_{ij}$ prediction and the same
closed-triangle rule as any ordinary face. This keeps a placeholder for the
feature in the octree hierarchy so that $\mathbb{S}$ still has an ancestor to
refine at the next level, instead of silently erasing structure the current
resolution cannot yet resolve into a proper triangle.

\subsection{Training and Inference Algorithms}
\label{sec:appendix-algorithms}
We provide pseudocode for the mask-uniform training step and the
corresponding mask$\to$token$\to$token inference loop of the mask-uniform
noise strategy in Sec.~\ref{sec:model}.

\begin{algorithm}[h]
\caption{Mask-uniform training step}
\label{alg:train}
\textbf{Input:} ground-truth slots $z^\star$ at level $l$, point cloud
$\mathbf{c}$, mask probability $p$, self-generation probability
$p_{\text{selfgen}}$, condition-drop probability $p_{\text{drop}}$, focal
exponent $\gamma$, confidence loss weight $\lambda$, wrong-class weight
$\kappa$
\begin{enumerate}
    \item $\mathbf{c} \leftarrow \varnothing$ w.p.\ $p_{\text{drop}}$ \hfill{\small\textit{(classifier-free guidance dropout)}}
    \item sample $t\sim\mathcal{U}(0,1)$; \ $b_{f,k}\overset{\text{iid}}{\sim}\mathrm{Bernoulli}(r(t))$ for every slot \hfill{\small\textit{(cosine schedule)}}
    \item sample $s\sim\mathrm{Bernoulli}(p)$
    \item \textbf{if} $s=1$ \textbf{(mask pathway):}
    \begin{enumerate}
        \item $\tilde z_{f,k} \leftarrow \mathtt{[MASK]}$ if $b_{f,k}{=}1$ else $z^\star_{f,k}$, for every slot
        \item $h \leftarrow f_\theta(\tilde z, t, l, \mathbf{c};\, \mathrm{mode}{=}1)$
        \item \textbf{return} $\mathcal{L}_{\text{mask}} = \mathcal{L}_{\mathrm{CE}}(\{(f,k):b_{f,k}{=}1\})$ \hfill{\small\textit{(Eq.~\ref{eq:focal}, focal CE on corrupted slots only)}}
    \end{enumerate}
    \item \textbf{else (uniform pathway):}
    \begin{enumerate}
        \item \textbf{if} $\mathrm{rand}() < p_{\text{selfgen}}$:
        \begin{enumerate}
            \item sample fresh $t'\sim\mathcal{U}(0,1)$, $b'_{f,k}\sim\mathrm{Bernoulli}(r(t'))$
            \item $z^{\text{mask}}_{f,k} \leftarrow \mathtt{[MASK]}$ if $b'_{f,k}{=}1$ else $z^\star_{f,k}$
            \item $\tilde z \leftarrow \arg\max\, p_\theta\big(\cdot \mid f_\theta(z^{\text{mask}},t',l,\mathbf{c};\mathrm{mode}{=}1)\big)$ \hfill{\small\textit{(no-grad self-generated corruption)}}
        \end{enumerate}
        \item \textbf{else:} $\tilde z_{f,k} \leftarrow u_{f,k}\sim\mathcal{U}\{0,\dots,255\}^{K_k}$ if $b_{f,k}{=}1$ else $z^\star_{f,k}$
        \item $h \leftarrow f_\theta(\tilde z, t, l, \mathbf{c};\, \mathrm{mode}{=}0)$
        \item $y_{f,k} \leftarrow \mathbb{1}[\arg\max\, p_\theta(\cdot\mid h_{f,k}) = z^\star_{f,k}]$ for every slot
        \item $\mathcal{L}_{\text{conf}} \leftarrow \frac{1}{9n}\sum_{f,k}\kappa_{f,k}\,\mathrm{BCE}\big(g_\phi(\mathrm{sg}[h_{f,k}]), y_{f,k}\big)$ \hfill{\small\textit{(Eq.~\ref{eq:lconf})}}
        \item \textbf{return} $\mathcal{L}_{\text{unif}} + \lambda\,\mathcal{L}_{\text{conf}}$, with $\mathcal{L}_{\text{unif}}=\mathcal{L}_{\mathrm{CE}}(\text{all slots})$
    \end{enumerate}
\end{enumerate}
\end{algorithm}

\begin{algorithm}[h]
\caption{Mask$\to$token$\to$token inference}
\label{alg:infer}
\textbf{Input:} parent conditioning, level $l$, point cloud $\mathbf{c}$,
steps $T$, re-mask schedule $\mathrm{sched}(s)$ linearly annealed
$0.2\!\to\!0.9$
\begin{enumerate}
    \item $z_{f,k} \leftarrow \mathtt{[MASK]}$ for every slot \hfill{\small\textit{(fully corrupted canvas)}}
    \item \textbf{for} $s = 0, \dots, T-1$:
    \begin{enumerate}
        \item $t \leftarrow s / T$
        \item $h \leftarrow f_\theta(z, t, l, \mathbf{c};\, \mathrm{mode}{=}1)$
        \item $z_{f,k} \leftarrow \arg\max\, p_\theta(\cdot \mid h_{f,k})$ for every slot \hfill{\small\textit{(mask-fill: overwrites all slots, no freezing)}}
        \item $h \leftarrow f_\theta(z, t, l, \mathbf{c};\, \mathrm{mode}{=}0)$
        \item $z_{f,k} \leftarrow \arg\max\, p_\theta(\cdot \mid h_{f,k})$ for every slot \hfill{\small\textit{(uniform-refine, reusing $h$ below)}}
        \item \textbf{if} $s < T-1$:
        \begin{enumerate}
            \item $\mathrm{conf}_{f,k} \leftarrow \sigma(g_\phi(h_{f,k}))$ for every slot
            \item $z_{f,k} \leftarrow \mathtt{[MASK]}$ w.p.\ $(1-\mathrm{conf}_{f,k})\cdot\mathrm{sched}(s)$ 
        \end{enumerate}
    \end{enumerate}
    \item \textbf{return} $z$
\end{enumerate}
\end{algorithm}

Algorithm~\ref{alg:train} details one training step of the mask-uniform
noise strategy of Sec.~\ref{sec:model}, and Algorithm~\ref{alg:infer} the
corresponding mask$\to$token$\to$token decoding loop used at generation
time. Both are written for a single sample; in practice all steps are
batched over the variable-length face sequence via the varlen attention
packing described in Sec.~\ref{sec:model}. Two details are elided from the pseudocode
for brevity: (i) classifier-free guidance, when enabled, evaluates every
$f_\theta$ call twice (with $\mathbf{c}$ and with $\mathbf{c}{=}\varnothing$)
and linearly interpolates the two logit sets before the $\arg\max$; (ii) the
$\arg\max$ over a connectivity slot's 8-row distribution is not a single
categorical draw but the autoregressive roll-out of Eq.~\ref{eq:matrixhead},
decoding one row at a time conditioned on the rows already emitted.

\subsection{Geometry-Anchored RoPE Anchor Table}

Unlike TSSR's 1D sequence indexing, we represent meshes via orderless per-face token blocks. To provide spatial and topological awareness, we condition rotary position embeddings (RoPE) on each token's 3D voxel coordinates and its intra-face role $\tau$ (vertex occupancy, intra-parent, or inter-parent connectivity), maintaining permutation equivariance across face blocks. For RoPE embedding dimension $64$, we allocate 2 frequency pairs to $\tau$ and 30 to geometry.
Each anchor is an explicit vertex (Table~\ref{tab:anchors}). At the coordinate level of the hourglass transformer, tokens are anchored to an ordered pair $(v_a, v_b)$: vertex-related tokens carry $(v_a, v_a)$ and connection tokens carry $(v_a, v_b)$. Sharing rotation angles across half their geometric frequencies turns vertex-edge incidence into an explicit phase alignment rather than an inferred latent relation. Sorting parent vertices lexicographically z-y-x ensures canonical edge pairs across shared faces. After token reduction, each coarse token remains anchored across all three parent vertices.

\label{sec:appendix-rope}
Table~\ref{tab:anchors} lists, for every token type at every hourglass
scale, the explicit vertex (or vertex pair) it is anchored to and its role
$\tau$, referenced from Sec.~\ref{sec:rope}.

\begin{table}[h]
\centering
\caption{Geometry-anchored rotary encoding. The $30$ geometric slots are split as $2$ anchors
$\times\ 3$ axes $\times\ 5$ slots at $9/$face, and as $3$ vertices $\times\ 3$ axes $\times\ 3$
slots after reduction. $\tau$ separates tokens that share an anchor.}
\label{tab:anchors}
\small
\begin{tabular*}{\textwidth}{@{\extracolsep{\fill}}lllc@{}}
\toprule
Hourglass level & Token Corresponds to & Anchor & Type $\tau$ \\
\midrule
\multirow{3}{*}{$9/$face, coordinate-level}
& $F_0,F_1,F_2$
& $(v_0,v_0),(v_1,v_1),(v_2,v_2)$
& $0$ \\
& $\Mintra_0,\Mintra_1,\Mintra_2$
& $(v_0,v_0),(v_1,v_1),(v_2,v_2)$
& $1$ \\
& $\Minter_{01},\Minter_{02},\Minter_{12}$
& $(v_0,v_1),(v_0,v_2),(v_1,v_2)$
& $2$ \\
\midrule
$3/$face, vertex-level& subdivision / intra / inter aggregate & $(v_0,v_1,v_2)$ & $0/1/2$ \\
$1/$face, face-level& face feature & $(v_0,v_1,v_2)$ & $3$ \\
\bottomrule
\end{tabular*}
\end{table}

\subsection{Focal Loss for Connectivity Class Imbalance}
\label{sec:appendix-focal}
Both $\mathcal{L}_{\text{mask}}$ and $\mathcal{L}_{\text{unif}}$ (Sec.~\ref{sec:model}) replace
the plain cross-entropy of Eq.~\ref{eq:simple-loss} with a multi-class focal
loss~\citep{Linfocal2017}. With $p_{f,k}$ the softmax probability assigned to the
ground-truth class $z^\star_{f,k}$,
\begin{equation}
    -\log p_\theta(z^\star_{f,k}\mid\cdot)
    \ \longrightarrow\
    -(1-p_{f,k})^{\gamma}\log p_{f,k},
    \label{eq:focal}
\end{equation}
which automatically down-weights slots the model already predicts confidently ($\gamma{=}0$
recovers plain cross-entropy) and keeps the $\sim\!92\%$ empty rows of the connectivity matrix
$\mathbf{M}$ from dominating the gradient.

\subsection{Denoising Within a Resolution Level}
\label{sec:appendix-inner-step}

\begin{figure}[h]
\centering
\includegraphics[width=\textwidth]{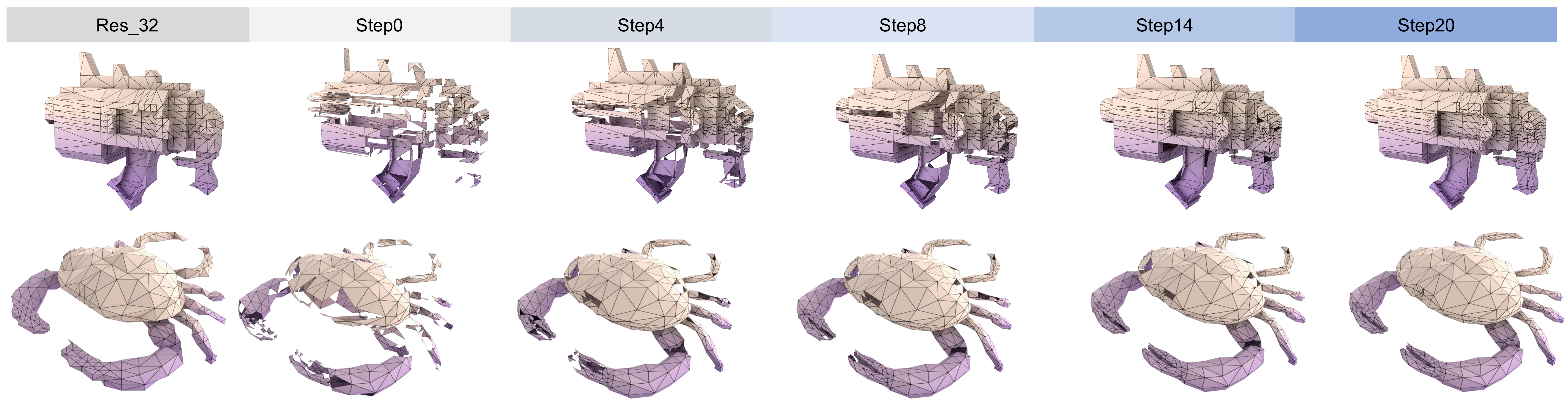}
\caption{\textbf{Intra-scale denoising.}
The left column shows the resolution-32 parent mesh. The remaining columns show the decoded predictions immediately before re-masking at diffusion steps 0, 4, 8, 14, and 20 of the same resolution transition.}
\label{fig:inner-step}
\end{figure}

Figure~\ref{fig:inner-step} visualizes intra-scale denoising within a single resolution transition, starting from the resolution-32 parent mesh. At each selected diffusion step, we decode the current prediction immediately before low-confidence tokens are re-masked for the next iteration. The initial prediction contains fragmented and disconnected geometry. As denoising proceeds, iterative prediction and re-masking progressively improve the object silhouette and local connectivity, leading to a coherent mesh at the final step.

\subsection{Generation Across Resolutions}
\label{sec:appendix-resolution-progression}

\begin{figure}[h]
\centering
\includegraphics[width=\textwidth]{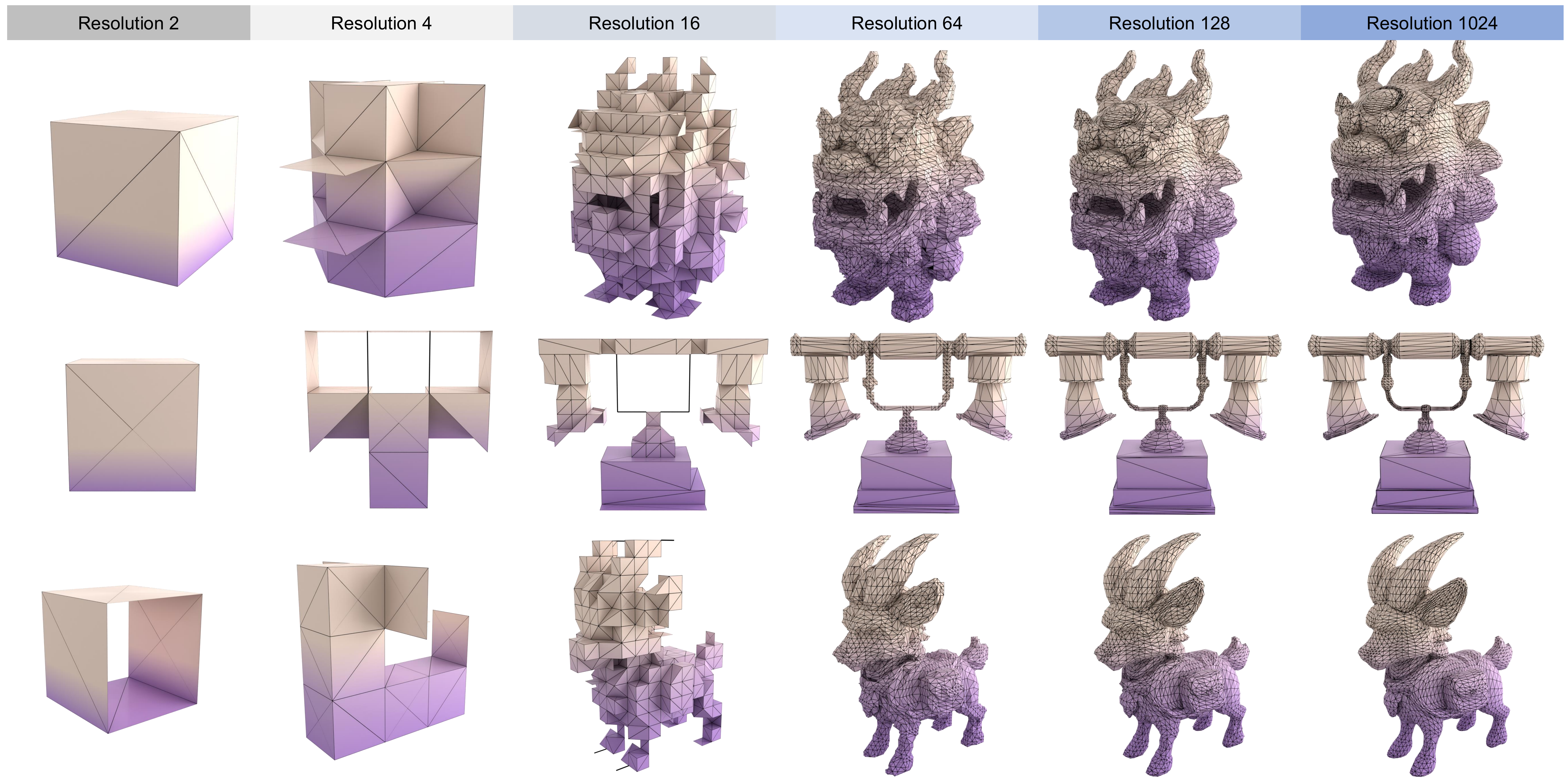}
\caption{\textbf{Mesh generation across spatial resolutions.}
Intermediate meshes from one coarse-to-fine generation trajectory.
Each column represents the completed mesh at a different grid
resolution. Increasing resolution progressively adds geometric detail
and refines connectivity; the result at each level can be used as a
standalone level-of-detail mesh.}
\label{fig:resolution-progression}
\end{figure}

Figure~\ref{fig:resolution-progression} illustrates the coarse-to-fine
generation trajectory across successive spatial resolutions. At lower
resolutions, the mesh captures the overall silhouette and the arrangement
of major parts with relatively few vertices and faces. Each subsequent
split-and-rewire transition doubles the coordinate-grid resolution,
introducing finer geometric features and more detailed connectivity while
building on the mesh produced at the preceding level. The resulting
meshes form a sequence of usable levels of detail: generation may stop
at an intermediate resolution when a coarser mesh is sufficient.

\subsection{Inference Speed Comparison}
\label{app:speed}
\begin{figure}[h]
    \centering
    \includegraphics[width=0.85\linewidth]{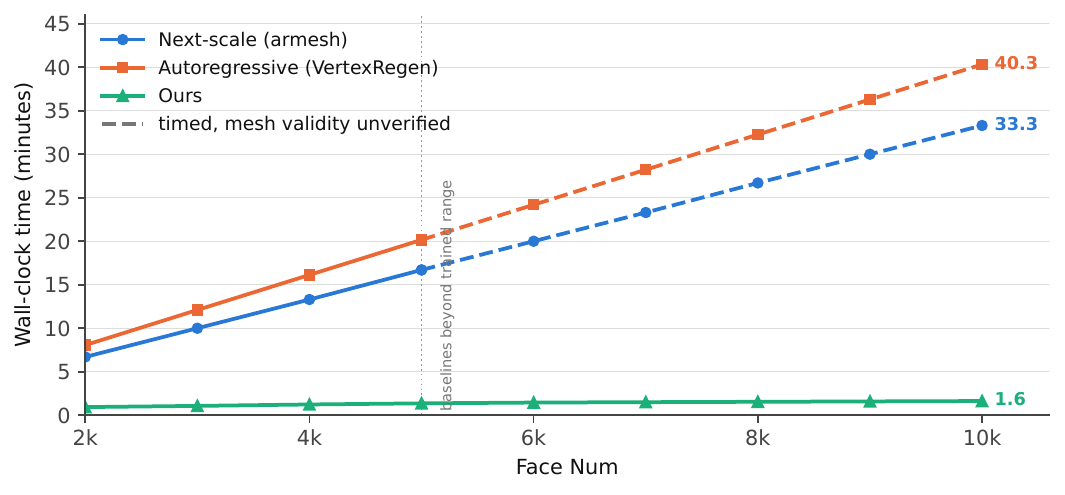}
    \caption{Inference wall-clock time versus output mesh complexity
    (number of faces), for the next-scale baseline (armesh), the
    autoregressive vertex-split baseline (VertexRegen), and \method{}.
    \method{} generates comparably complex meshes over an order of
    magnitude faster than both baselines. All curves are directly
    measured. Neither baseline was trained beyond $5\text{k}$ faces
    (dotted vertical line); their dashed segments are still measured
    wall-clock time, but the decoded mesh's validity at that scale is
    unverified. The \method{} curve is solid throughout: every bin is a
    directly measured, verified generation.}
    \label{fig:speed_comparison}
\end{figure}

We measure wall-clock inference time as a function of output mesh
complexity (number of faces in the generated mesh) against a next-scale
autoregressive baseline (armesh) and a token-by-token autoregressive
vertex-split baseline (VertexRegen~\cite{zhang2025vertexregen}), as well
as \method{}. All methods were run on a single GPU. Neither baseline was trained on meshes beyond
$5\text{k}$ faces; for the $[5\text{k}, 10\text{k}]$ range (dashed in
Figure~\ref{fig:speed_comparison}) we still measure the wall-clock time
each baseline takes to reach that many faces, but do not verify that the
resulting mesh is valid at that scale -- these points reflect real,
measured decoding cost, not usable output quality. For \method{}, every
point across $[2\text{k}, 10\text{k}]$ is likewise a directly measured
average wall-clock time over batches of generated trellis test-set
meshes with real samples in every bin, with the decoded mesh verified
at every point.

\end{document}